\documentclass{article}

\usepackage{automl2019}

\usepackage{subfigure}
\usepackage{wrapfig}
\usepackage[utf8]{inputenc} 
\usepackage[T1]{fontenc}    
\usepackage{hyperref}       
\usepackage{url}            
\usepackage{booktabs}       
\usepackage{amsfonts}       
\usepackage{nicefrac}       
\usepackage{microtype}      
\usepackage{color}
\usepackage{graphicx}
\usepackage{xspace}
\usepackage{amsmath}
\usepackage{textcomp}
\usepackage{amsmath}

\DeclareMathOperator*{\argmin}{arg\,min}

\usepackage{hyperref}

\makeatletter
\newcommand{\printfnsymbol}[1]{%
  \textsuperscript{\@fnsymbol{#1}}%
}
\makeatother

\newcommand{\name}{NP\xspace}

\title{Neural Architecture Search Over a Graph Search Space}

\jmlrheading{S. Jastrzębski, Q. de Laroussilhe, M. Tan, X. Ma, N. Houlsby and A. Gesmundo}

\ShortHeadings{Neural Architecture Search Over a Graph Search Space}{Jastrz\k{e}bski, de Laroussilhe, Tan, Ma, Houlsby and Gesmundo}
\firstpageno{1}

\author{\name Stanisław Jastrzębski\thanks{Work done during an internship at Google} \email staszek.jastrzebski@gmail.com\\
       \addr Jagiellonian University
       \vskip-0.09in
       \AND
       \name Quentin de Laroussilhe \email underflow@google.com \\
       \vskip-0.33in
       \AND
       \name Mingxing Tan \email tanmingxing@google.com \\
       \vskip-0.33in
       \AND
       \name Xiao Ma \email xima@google.com \\
       \vskip-0.33in
       \AND
       \name Neil Houlsby \email neilhoulsby@google.com\\
       \vskip-0.35in
       \AND
       \name Andrea Gesmundo \email agesmundo@google.com\\
       \addr{Google AI}
       }

\begin{document}


\maketitle

\begin{abstract}
Neural Architecture Search (NAS) enabled the discovery of state-of-the-art architectures in many domains.
However, the success of NAS depends on the definition of the search space. 
Current search spaces are defined as a static sequence of decisions and a set of available actions for each decision.
Each possible sequence of actions defines an architecture.
We propose a more expressive class of search space: directed graphs.
In our formalism, each decision is a vertex and each action is an edge.
This allows us to model iterative and branching architecture design decisions.
We demonstrate in simulation, and on image classification experiments, basic iterative and branching search structures, and show that the graph representation improves sample efficiency.
\end{abstract}

\section{Introduction}

Neural network design requires many decisions, involving human expertise and time. AutoML speeds up this process by automating some of these decisions~\citep{DBLP:journals/corr/abs-1810-13306}.
%
Neural Architecture Search (NAS), as used in ~\citep{zoph2017}, is an approach to AutoML that uses a neural network (``the controller'') to explore different architectures and find the best.
The parameters of the controller's Neural Network are optimized to maximize the performance of generated networks on the downstream task.
This approach has achieved many recent successes \citep{zoph2017b,ramachandran2018searching,bello2017neural} 


The success of NAS hinges on the definition of the search space of architectures. 
Standard search spaces are defined as a linear chain of decisions and a set of available actions for each decision.
For example, the first decision might be ``select the learning rate'' and the corresponding set of actions might be: $\{0.1, 0.01\}$, and the second decision might be ``choose the number of layers'' and its set of available actions might be: $\{1, 2, 3\}$.
The controller samples one architecture by sampling one action for each decision. 
However, this formalism hinders the application of NAS to problems that are difficult to define as a linear sequence of actions.

For example, consider a search space that first picks between more optimizers and then chooses the hyperparameters for the picked optimizer. Performing search over this space using a static sequence forces the controller to make decisions about hyperparameters of all optimizers, rather than only of the picked one. Or as another example, consider a search space that stacks an arbitrary number of convolutional layers and choose different parameters for every added convolution, with a linear search space this iterative component can be represented only by rolling out the loop for a fixed number of iterations.

The main goal of this paper is to lift the limitation of the linear search space for NAS methods. 
We construct the search spaces as a graph, where each decision is a vertex and each action is an edge.
Thus the sequence of decisions defining an architecture is not fixed but is determined dynamically by the choices of actions.
With this approach the controller may branch and iterate over states.
%
%
To perform NAS over the proposed graph search space, we design a novel neural controller, capable of dynamically selecting a path through the graph and condition the actions' distributions on the the prior actions.



\section{Neural architecture search as a walk in a graph}

NAS aims to find the optimal architecture for a given ML task.
To do this, NAS takes an iterative approach.
At each iteration, a generative neural network, called the \emph{controller}, proposes a new architecture.
Next, this architecture is trained and evaluated on the ML task.
Finally, the evaluation metric is used as a reward to update the controller, using an RL algorithm.

\citet{zoph2017} propose to model the search space as a fixed length sequence of states $s_{1:T}$: one state for each decision.
This approach allows to implement the controller as an RNN.
In each state $s_t$ the controller takes one of the available actions $a_t$. The final sequence of actions $a_{1:T}$ defines the new architecture to be trained and evaluated.

Modeling the search space as a static length sequence has important limitations. For example, consider a search space that first selects an optimizer and then chooses the hyperparameters for the selected optimizer. Performing search over this space using a static sequence of decision forces the agent to make decisions about hyperparameters of all optimizers, rather than only the selected one.
This is an example of a search space branching.
In general, a linear search space cannot efficiently model either branching or iterative search spaces.

To overcome these limitations, we propose to model the search space as a directed graph, 
$\mathcal{G}=(V,A)$,
where each state is represented by a vertex,  $v_{t} \in V$,
and each action is represented by a directed edge, $a_{t} \in A $. 
Note that this formalism allows for multiple edges between each pair of vertices, and the graph can be cyclic.
Each path from the start state to a terminal state yields a sequence of actions defining an architecture.

When traversing the graph, paths are sampled, starting from the start state $v_0$.
Actions are sampled from the distributions learned by the controller $p_{v_t}(a_t)$. The next vertex, $v_{t+1}$, is determined by following the sampled edge $a_t$.
The walk is terminated upon reaching any of the terminal vertices.
After termination, the sampled architecture is defined by the sequence of actions $(a_1,\ldots,a_T)$.
Note that $T$ does not have to be constant, and each state can be visited multiple times. 


\section{Dynamic Neural Controller}
\label{sec:arch}


\begin{figure}[t]
\begin{center}
\includegraphics[width=0.7\linewidth]{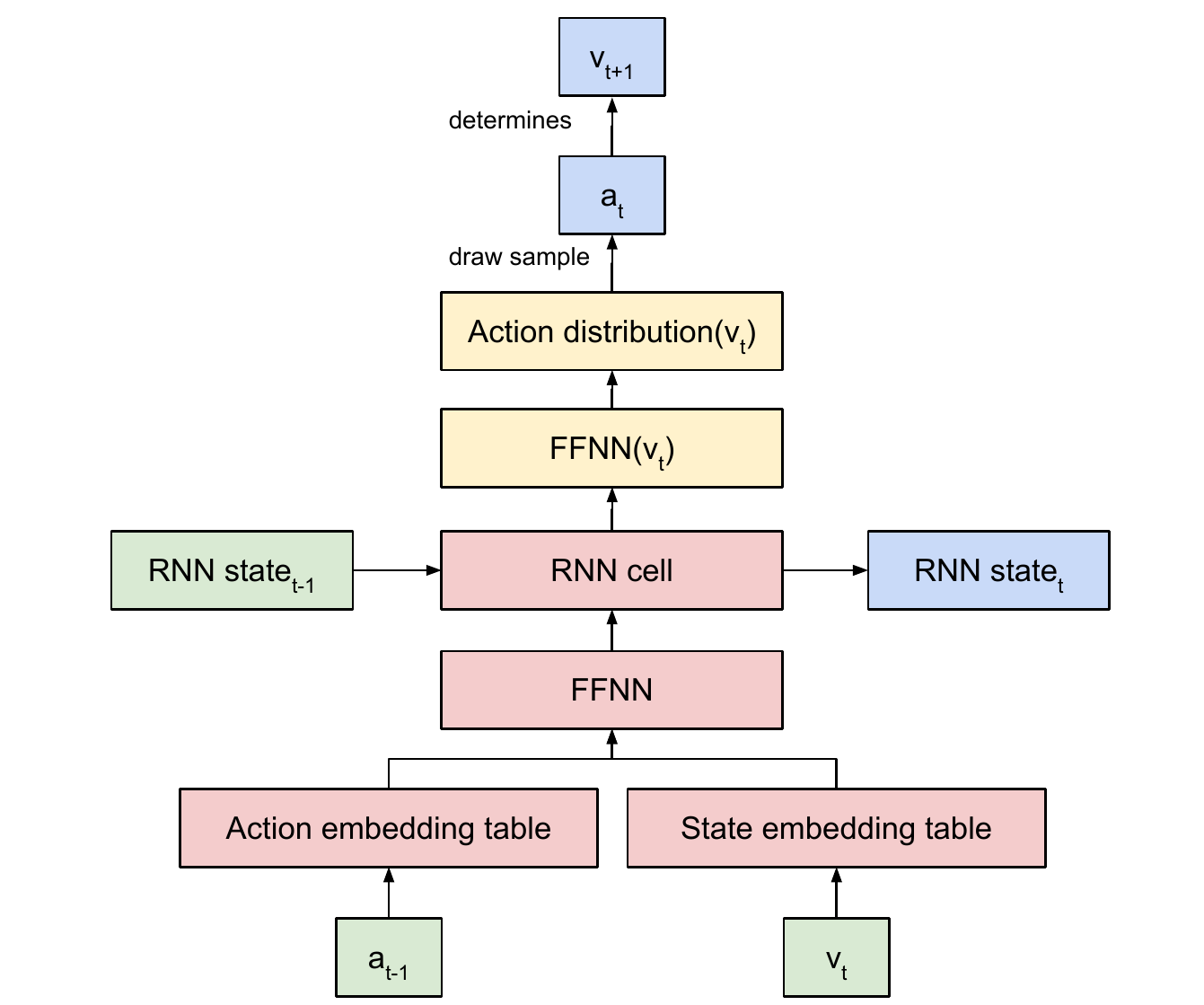}
\caption{
Architecture for one timestep of the dynamic controller. Inputs are highlighted in green, outputs in blue. The timestep-independent network is highlighted in red, the timestep-dependent network in yellow.
\label{fig:architecture}
}
\end{center}
\end{figure}

To optimize over a graph search space, we propose a modification of the RNN-based neural controller model from~\citet{zoph2017}.
The proposed architecture allows the controller to sample a path from an arbitrary graph, while dynamically setting the sequence of RNN steps to match the sequence of sampled states.
A dynamic architecture is needed to explore graph search spaces, since each action distribution is inferred by a distinct portion of the controller architecture,
and the action distribution to sample from at a given timestep $t$ is determined only after prior timestep $t-1$ is executed.
Instead, for the linear search space, each RNN timestep $t$ always correspond the same action distribution, thus allowing to work with a static controller architecture.

Figure~\ref{fig:architecture} depicts the architecture for one timestep of the dynamic RNN controller.
Inputs are highlighted in green, outputs in blue. The timestep-independent network is highlighted in red, the timestep-dependent network in yellow.
At each timestep, the controller takes as input: 1) the current state $v_t$, 2) the previous action $a_{t-1}$, 3) the RNN cell state.
These inputs are aggregated by the timestep independent network.
This is the part of the network whose parameters are the same at each timestep, which includes the action and state embedding tables.
The embeddings of $a_{t-1}$ and $v_t$ are aggregated with a fully connected network. 
Then, this aggregation is the input to an LSTM RNN cell.
The following portion of the architecture is timestep dependent.
It has the function of inferring the action distribution of the current state $v_t$.
The outputs are: 1) the action $a_t$ sampled from the timestep dependent action distribution, 2) the next state $v_{t+1}$ determined by $a_t$ 3) the new RNN state.

The training is analogous to~\citep{zoph2017}.
The controller learns the action distributions with the objective of maximizing the expected reward of the generated architectures using policy gradient methods.

Notice that the dynamic controller is not required to visit all the states in the search space. This brings multiple benefits that might improve the sample efficiency and stability the controller. Firstly, the controller visits less states, making credit assignment easier. Secondly, gradient updates are performed only for actions that are relevant to the generated model.


\section{Experiments}

We demonstrate the potential benefit of performing architecture search with a graph search space, and the ability of the dynamic controller to optimize this space, using two artificial tasks.
We present hand-crafted optimization problems that are representative of the branching and iterative components of architecture design.
For each toy task, we compare the efficacy of modeling the search space as a sequence or a graph.

The first task models the iterative process of stacking layers.
The second task models a decision that involves branching, such as picking the optimizer among few alternatives, and tuning its parameters.
In both tasks we demonstrate that expressing the search space as a graph significantly improves the sample efficiency of the controller.


\subsection{Stack layers toy task}

\begin{wrapfigure}{r}{0.5\textwidth}
\begin{center}
\includegraphics[width=1.0\linewidth]{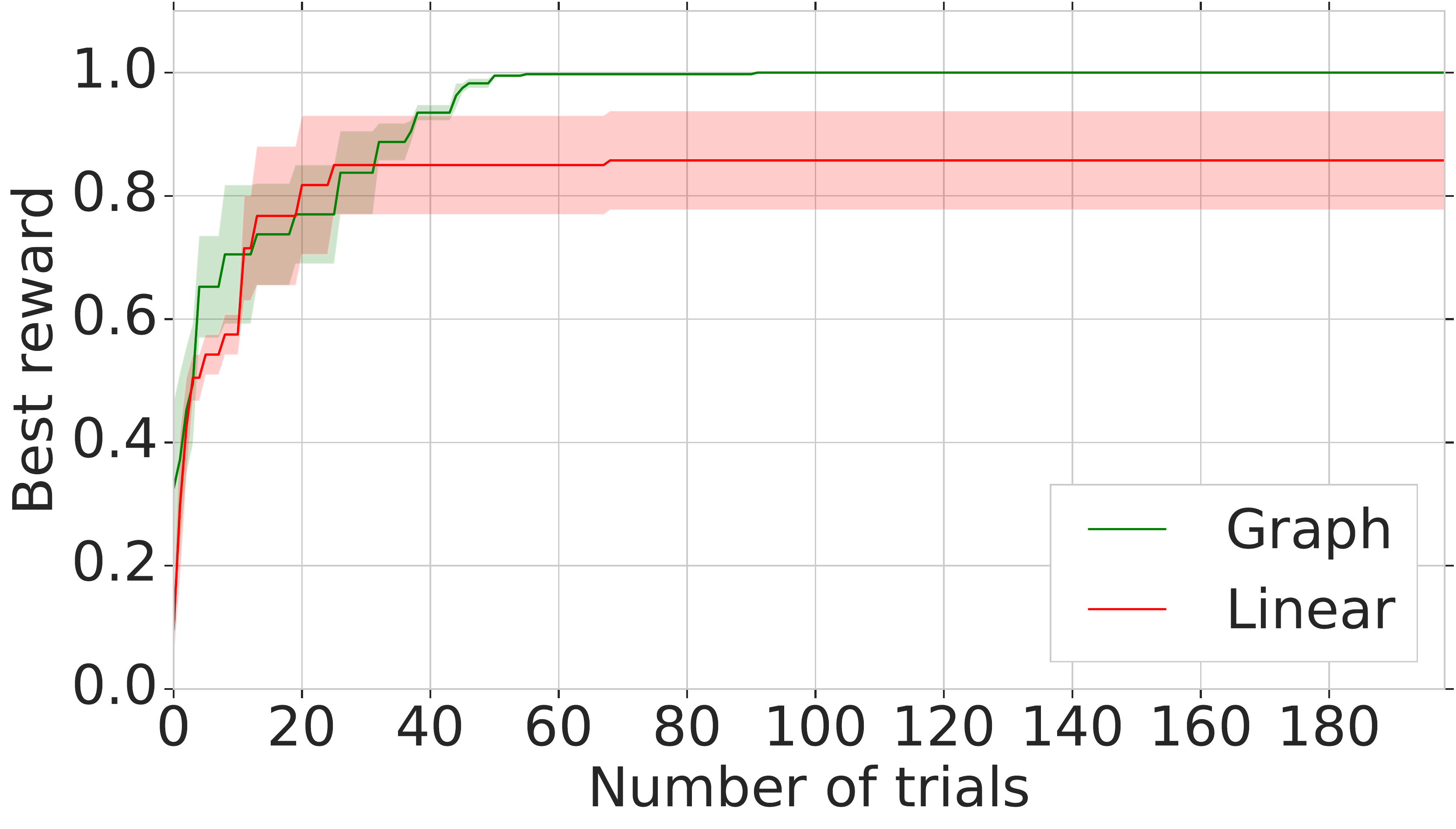}
\caption{Results for the ``stack layers'' task. Best reward over time averaged across 10 replicas. The shaded area corresponds to maximum and minimum reward achieved by the replicas. 
}
\label{fig:addlayers_results}
\end{center}
\end{wrapfigure}


The first task is based on an iterative, or cyclic, search space.
This task simulates the process of constructing a neural network with a variable number of layers.
The controller must choose the optimal number of layers $n$.
The reward in this simulation is defined as $-(n - L)^2$.
The maximum reward is achieved by constructing a network with exactly $L$ layers.
Otherwise, a squared error penalty is applied.

To compare the two approaches, we define both linear and graph instances of this search space.
A linear search space describing this task is defined as a sequence of states of length $2L$.
At each state only 2 actions are available, either to add 1 extra layer or to terminate.
The graph search space is defined as a single state with 2 actions.
The action to add a new layer is represented as a cyclic edge on the unique state.
The second action transitions to the terminal state, after which no more layers are added. 

For the following experiments, $L$ is set to $10$.
The reward is normalized to $[0,1]$ range. We use REINFORCE to optimize controller,
with learning rate 0.001 and entropy regularization 0.1.

The results are summarized in Figure~\ref{fig:addlayers_results}.
The graph search space improves training speed, converging to the maximum reward much more rapidly.
In fact, in this search space, linear fails to attain maximal reward in the number of trials ran, 200. In the linear case the controller is forced to output a fixed number of decisions, which, in this task, forces an arbitrary upper bound (for we chose $2L$ based on prior knowledge of the simulated task).
However, the graph search space has no bound on number of iterations,
thus, lifting the constraint on the maximum number of layers required by the linear version.
The graph search space could be harder to optimize, because it may accept an infinite number of trajectories.
However, despite the added complexity, the dynamic controller is able to solve the task more effectively searching over the graph. 


\subsection{Select an optimizer toy task}


\begin{figure}[t]
\centering
\subfigure[Search space as a sequence]{\includegraphics[width=0.48\linewidth]{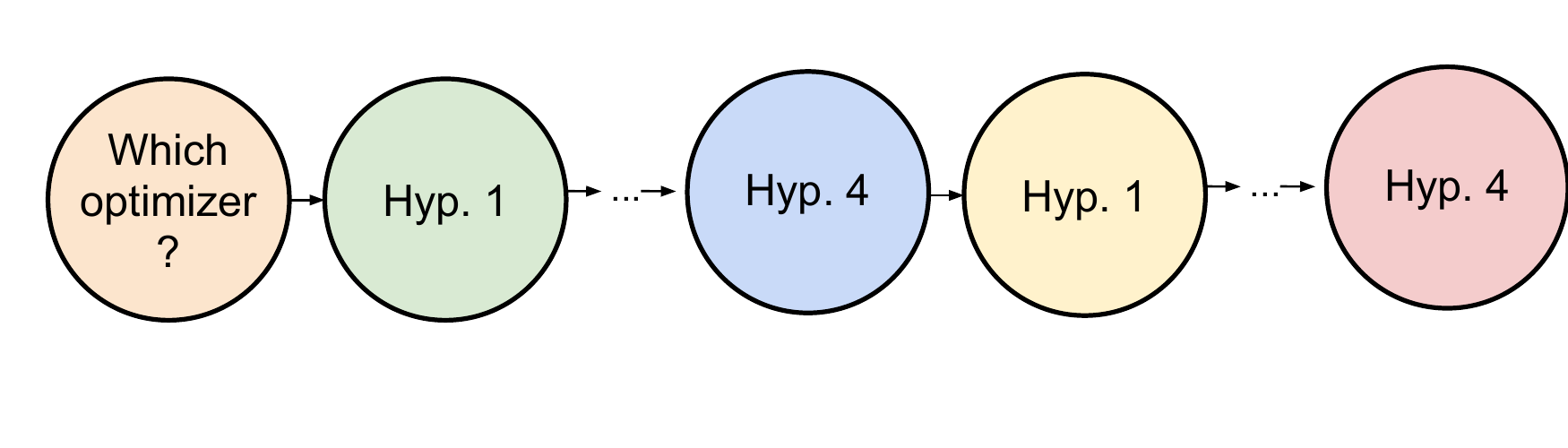}}
\subfigure[Search space as a graph]{\includegraphics[width=0.48\linewidth]{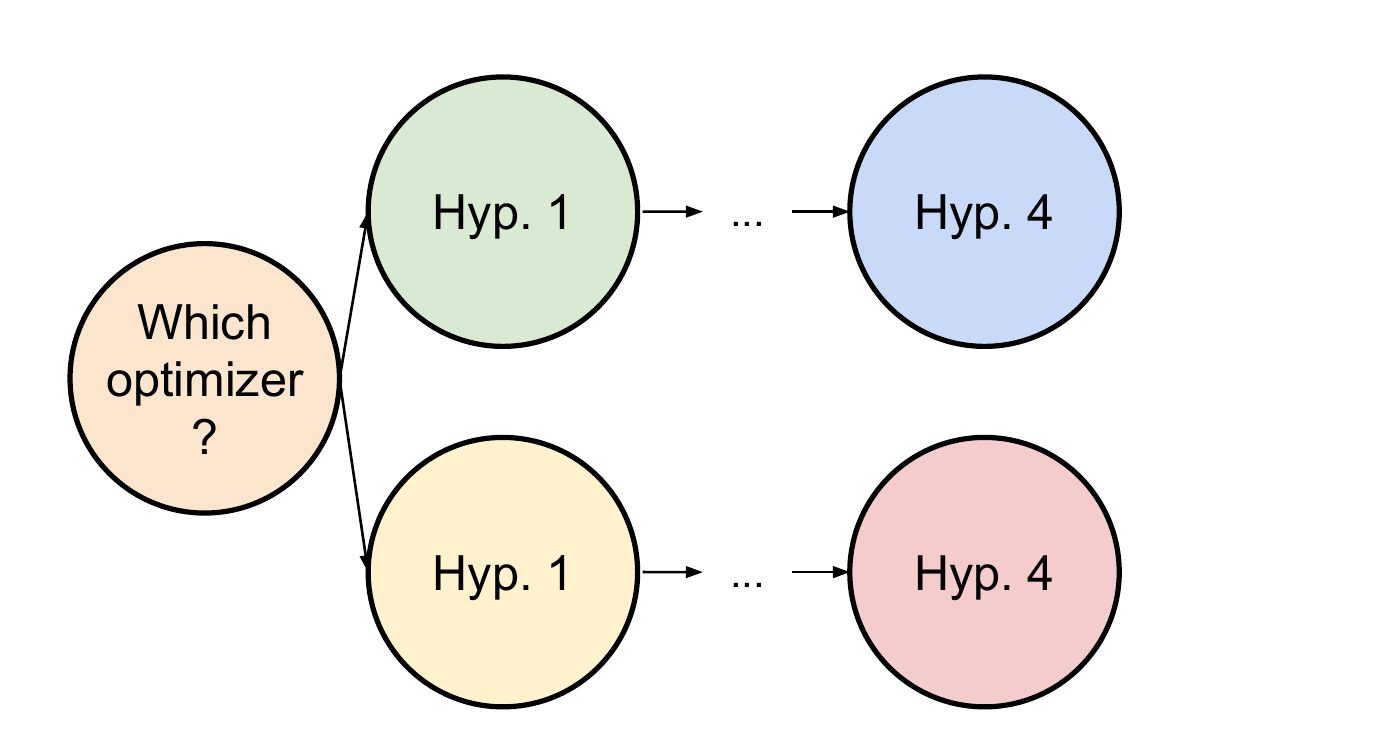}}
\caption{``Select an optimizer'' toy task with $B$ branches.}
\label{fig:branch_vis}
\end{figure}

\begin{figure}[t]
\begin{center}
\subfigure[Results for two branches environment.]{\includegraphics[width=0.49\linewidth]{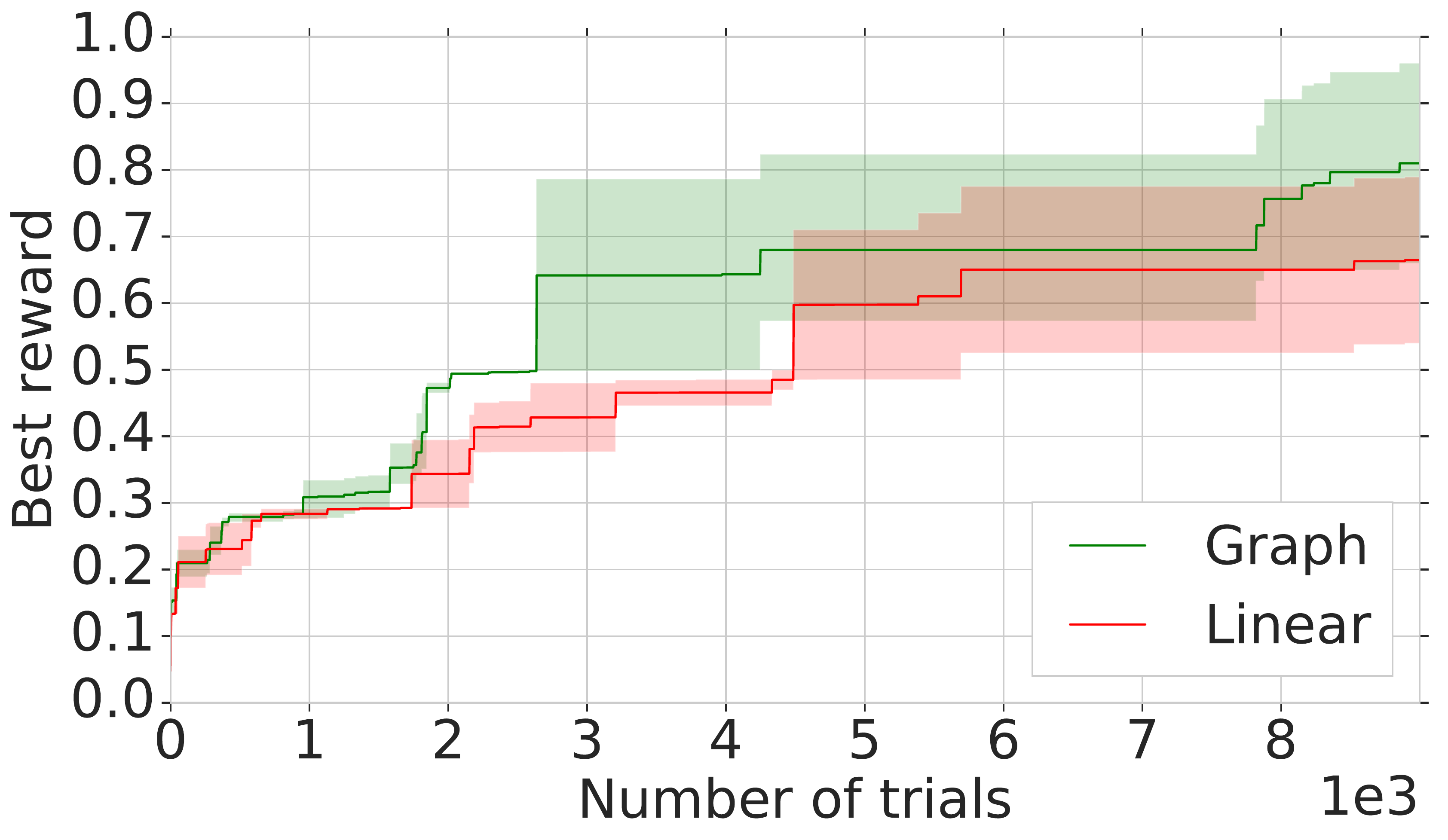}}
\subfigure[Results for four branches environment.]{\includegraphics[width=0.49\linewidth]{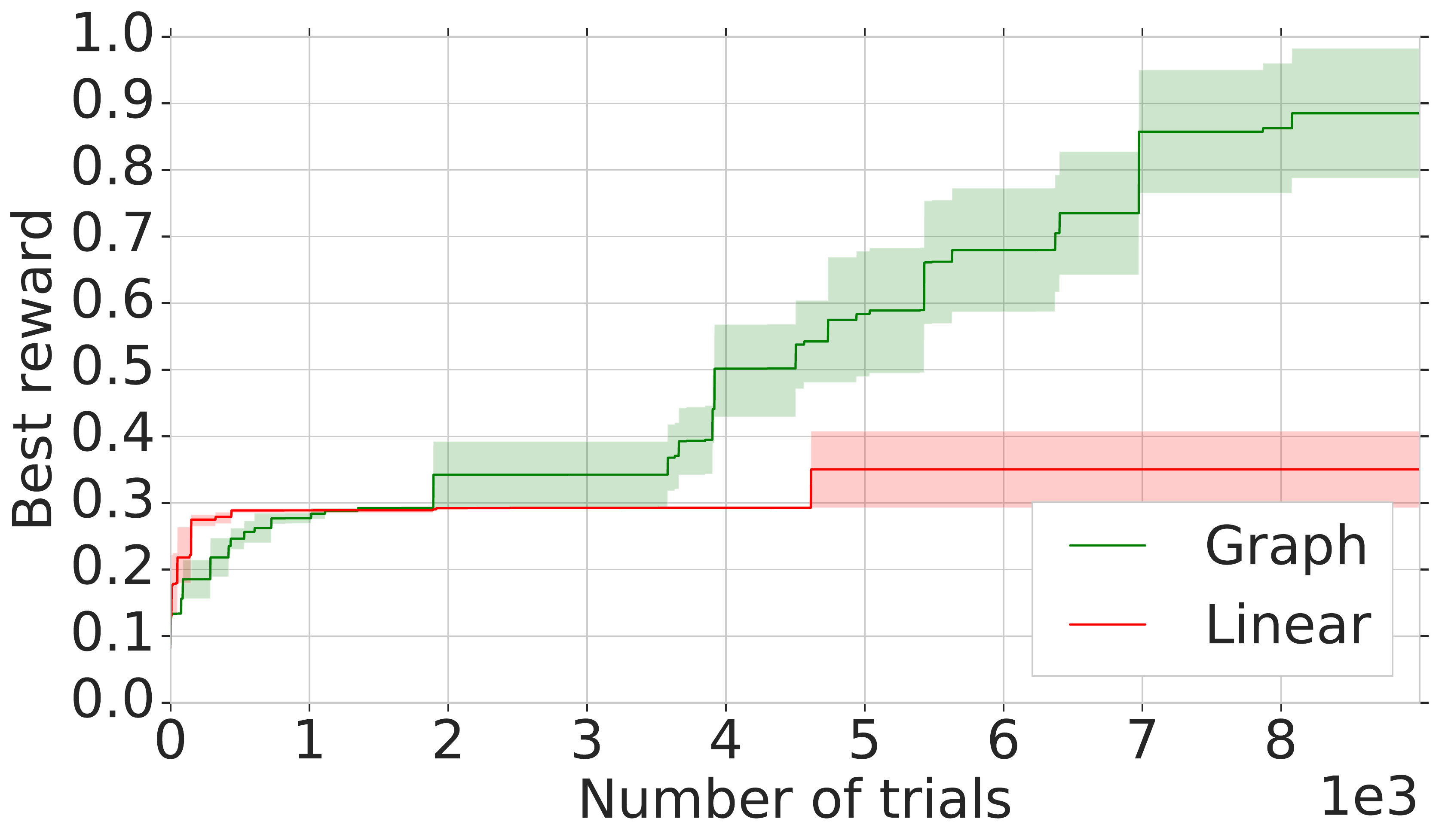}}
\caption{Results for the ``select an optimizer'' task. Best reward over time averaged across 10 replicas for two branch (left) and four branch (right) versions of the environment. The shaded area corresponds to maximum and minimum reward achieved by the replicas.
}
\label{fig:branch_results}
\end{center}
\end{figure}

The second task is based on an branching choices search space.
This task simulates picking between several optimizers, then choosing optimizer specific hyper-parameters, such as learning rate or decay rate.
Thus, we designed a toy task that requires the controller to first choose an optimizer $b$ among $B$ options, and then chooses 4 optimizer specific hyper-parameters: $p_1,p_2,p_3,p_4$.
For the following experiments $B$ may take a value in $\{2, 4\}$, and each of the hyperparameter can be set to a value in the range $[1, 100]$.

The reward is defined as follows. The agent is maximally rewarded for setting all the 4 relevant hyperparameters to the value of $50$.
Let  $M(b,p_1,p_2,p_3,p_4)=\argmin_{i=1,\ldots,4} p_i \neq 50$, i.e. $M$ denotes index of the first \emph{incorrect} $p_i$. Then, the unnormalized reward is defined as: $r = -(p_M - 50)^2 - \sum_{i=M+1}^4 50^2$.
We normalize the reward to $[0,1]$ range.
This choice of the reward function is motivated by our early experiments in which we found that random search can be a strong baseline. To highlight the potential of graph based search space we decided to use a reward which is difficult to optimize using a random search.

\begin{wrapfigure}{r}{0.5\textwidth}
\vskip -0.35in
\centering
\includegraphics[width=1.0\linewidth]{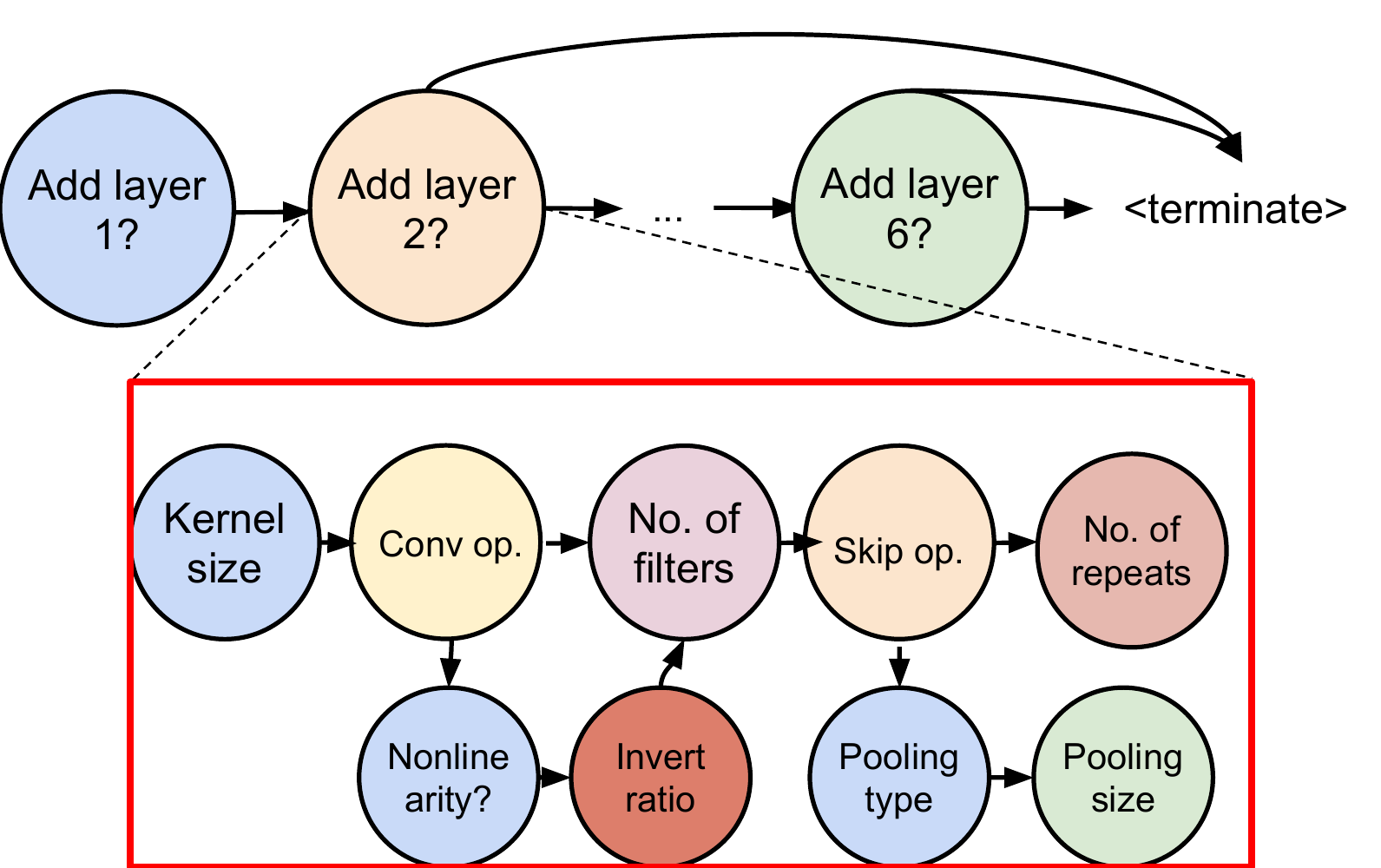}
\caption{Graph search space for Cifar10. 
}
\label{fig:nn_search_space}
\end{wrapfigure}

Figure~\ref{fig:branch_vis} illustrates how the ``select an optimizer'' task may be defined as linear and graph search space.
Both search spaces have one state to choose the optimizer, and then four states for each optimizer choice.
The main difference is that the graph search space allows one to build a tree structure having a branch dedicated to each optimizer specific list of hyperparameters.
The linear formulation requires flattening the tree structure into a sequence and choosing the hyperparameters of unused optimizers.

Again, the reward is normalized to $[0,1]$, and the controller is trained with REINFORCE with learning rate 0.0001 and entropy regularization 0.8.

The maximum reward achieved by each time step is shown in Figure~\ref{fig:addlayers_results} for environment with two branches (left) and four branches (right).
Similar to results on the previous toy task, the graph search space improves optimization speed.
We observe a larger benefit in the environment with more branches.

\subsection{Dynamic number of layers}

We experiment on an image classification task. As illustrated by the toy tasks, the key benefit of graph search space lies in the improved convergence speed. In particular, the controller updates only the weights corresponding to the states used to generate the current model.

\begin{figure}[t]
\begin{center}
\includegraphics[width=0.9\linewidth]{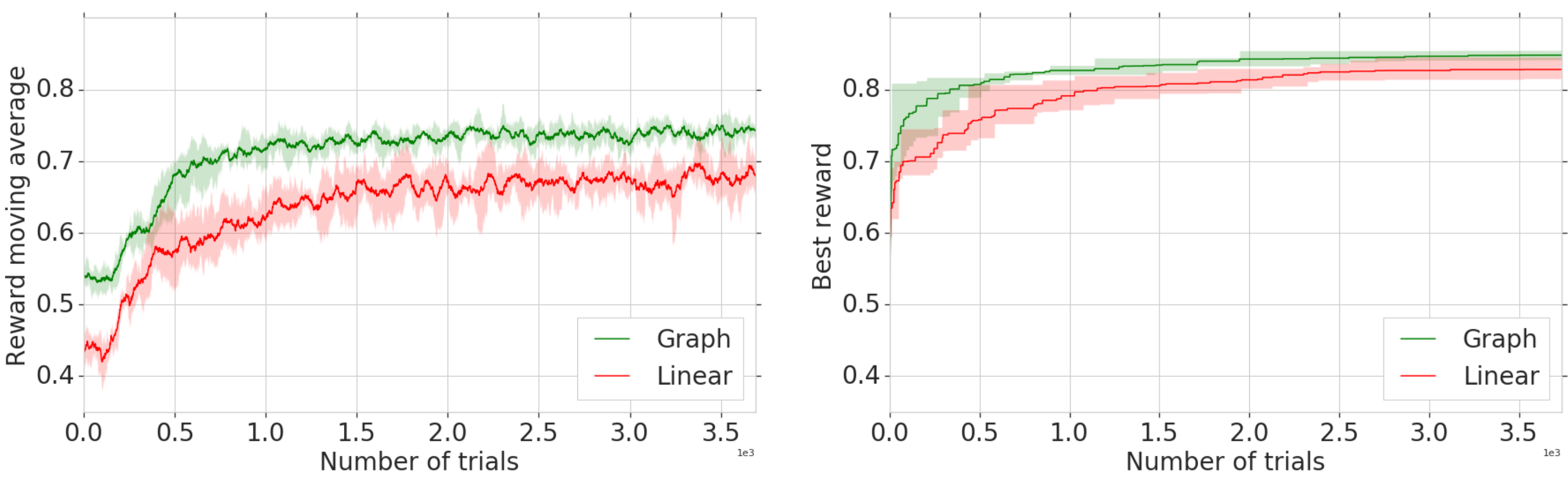}
\caption{Moving average (left) and best (right) reward averaged across 5 replicas for Cifar10. 
}
\label{fig:nn_learning_curves}
\end{center}
\end{figure}

We experiment on the Cifar10 dataset \citep{cifar-10} and use a similar search space to \citet{tan2018}. The key innovation of~\citet{tan2018} is using a \emph{hierarchical} search space, in which a each of the $7$ layers of the model can be defined individually.
To adapt the search space to Cifar10 we reduce the maximum number of filters, the number of repetitions of each layer, and the number of layers. Finally, we slightly extend the complexity of each layer by allowing the controller to pick the non-linearity and the expansion ratio in squeeze-and-excite layers. 
%
In contrast to the linear search space used in ~\citep{tan2018} we allow the controller to pick the number of layers. We append an extra decision after each final state of the layers whether or not to continue extending the network. The number of layers is capped at $6$. See also Figure ~\ref{fig:nn_search_space} for illustration.

Due to the higher resource requirement of this line of experiments we use the more sample efficient Priority Queue Training algorithm~\citep{abolafia2018neural}.
We pick learning rate $0.001$, and the entropy regularization $0.8$.
We sample $1500$ architectures in total. The generated models are trained for $20$ epochs instead of the standard $200$ epochs required to reach state of the art results on Cifar10.

\begin{figure}[t]
\begin{center}
\includegraphics[width=0.9\linewidth]{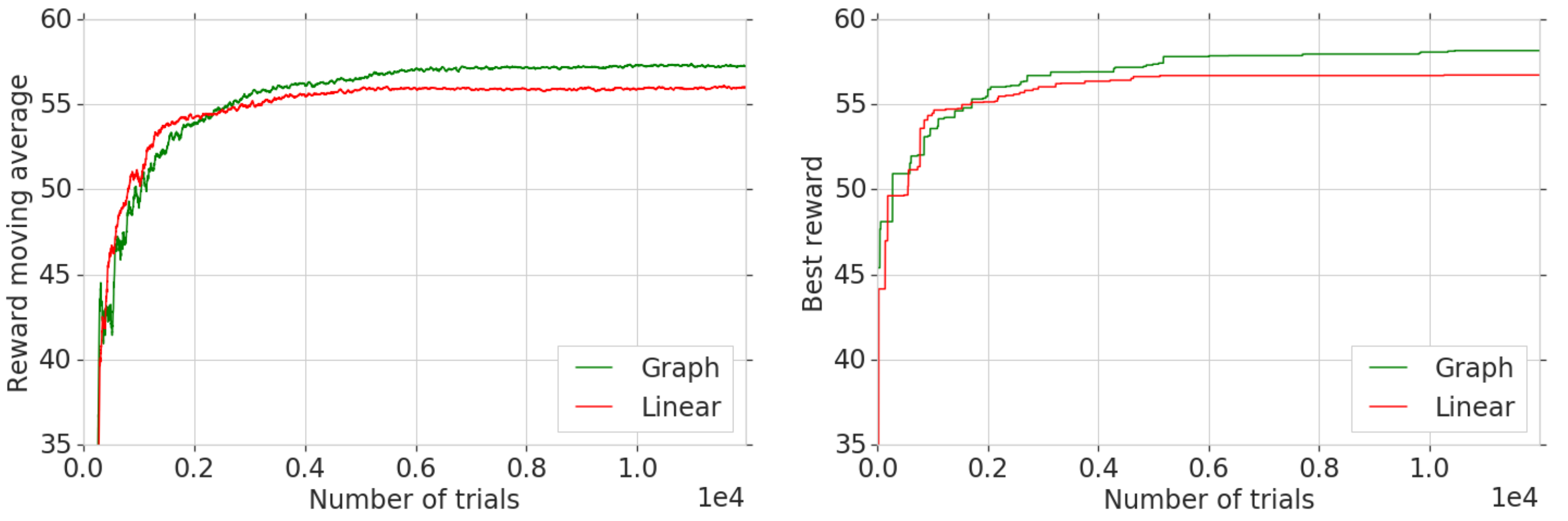}
\caption{Moving average (right) and best (left) reward for Imagenet. 
}
\label{fig:imagenetsearch}
\end{center}
\end{figure}


Results are summarized in Figure~\ref{fig:nn_learning_curves}. Using the graph search space leads to finding a better architecture, approximately $84\%$ accuracy on Cifar10 compared to $82\%$ of the best model generated using the linear search space.
Note that these results are not meant to reach state-of-the-art performance since the sampled models are trained on a significantly smaller training set and the models are much smaller than state-of-the-art models.
Avoiding training to convergence is the standard methodology for NAS experiments ~\citep{tan2018}. This approach allows to sample a higher number of models. Standard NAS methodology would require to select the best architectures found and train them on the full training set to be able to compare with state-of-the-art results, but this is beyond the scope of this work, that focuses on the analysis of the graph search space.




We have also evaluated the proposed graph search space on ImageNet dataset. Figure \ref{fig:imagenetsearch} shows the best reward obtained graph and linear search space. The graph search experiment converges faster and reaches higher accuracy than linear search space.

\section{Related Work}

The complexity of model engineering in machine learning is widely recognized.
Recent successes in automated model design have spurred further work in learning to learn~\citep{thrun2012learning}.
A variety of optimization methods have been proposed to search over architectures, hyperparameters, and learning algorithms.
These include random search \citep{bergstra2012random}, 
parameter modeling \citep{bergstra2013making}, 
meta-learned hyperparameter initialization \citep{feurer2015initializing},
deep-learning based tree searches over a predefined model-specification language \citep{negrinho2017deeparchitect},
learning of gradient descent optimizers \citep{wichrowska2017learned,bello2017neural},
and learning to generate network weights directly~\citep{ha2016hypernetworks,brock2017smash}.
An emerging body of neuro-evolution research has adapted genetic algorithms for these complex optimization problems \citep{conti2017improving}, including to set the parameters of existing deep networks \citep{such2017deep}, 
evolve image classifiers \citep{real2017large},
and evolve generic deep neural networks \citep{miikkulainen2017evolving}.

Our work is most closely related to deep RL based methods for auto ML.
Neural Architecture Search (NAS)~\citep{zoph2017} introduced the idea to use deep neural network controller, trained with RL, to generate architecture configurations.
NAS was applied to construct Convolutional Neural Networks (CNNs) for the CIFAR-10 task and Recurrent Neural Networks (RNNs) for the Penn Treebank tasks. 
Significant efforts have been made to achieve increased sample efficiency efficient using a progressive search procedure~\citet{liu2017progressive}, parameter sharing~\citep{cai2017reinforcement,pham2018efficient}, or transfer learning~\citep{wong2018transfer}.

This paper proposes a more expressive formalism for defining model architecture search space for RL based approaches, by generalizing the linear search spaces to graphs.
Extension of linear search spaces to tree structures have been proposed in prior work in the domain of automatic model selection and hyperparamter tuning for approaches based on Bayesian Optimization. These include using specific kernels  ~\citep{raiders13-workshop} or random forests ~\citep{thornton2013auto} to model the acquisition function.

While we focus on RL approaches to NAS, there is a large literature on alternative formulations of NAS. For instance, \citet{pham2018,liu2018darts} proposes to share parameters of all child networks, which allows for drastically faster convergence. However, this was shown to force using a rather limited search space~\citep{sciuto2019evaluating}.


\section{Conclusion}

We introduced a more expressive formalism for defining neural architecture search spaces, and a novel dynamic controller that is able to efficiently learn to explore this new class of search spaces.
Our experiments show significant improvements in sample efficiency in exploring common iterative or branching architecture design patterns.
As next steps, we plan to apply this method to other neural architecture search tasks and design a new class of search spaces that are not feasible in the linear form.

\bibliographystyle{unsrtnat}
{\small\bibliography{meta_learning}}




\end{document}